\documentstyle[eaclap,a4wide]{article}

\input{epsf}  %loads epsf.tex

\newenvironment{formula*}{\begin{displaymath}
\begin{array}{c}\displaystyle}{\end{array}\end{displaymath}}

\newcommand{\NUM}[1]{\refstepcounter{equation}
\label{#1}\'\indent(\theequation) }

\newcommand{\bdes}{\begin{description}}
\newcommand{\edes}{\end{description}}
\newcommand{\bite}{\begin{itemize}}
\newcommand{\eite}{\end{itemize}}
\newcommand{\benum}{\begin{enumerate}}
\newcommand{\eenum}{\end{enumerate}}

\newcommand{\sref}[1]{(\ref{#1})}

\font\tenss  = cmss10   % sans serif
\font\tenssi = cmssi10  % italic sans serif
\font\tenssbx = cmssbx10  % italic sans serif
\newcommand{\CL}[1]{{\let\em=\tenssi\let\bf=\tenssbx{\tenss #1}}\relax}

\date{}
\author{Janet Hitzeman, Marc Moens and Claire Grover\\
{\sc hcrc} Language Technology Group \\ The University of Edinburgh \\
2, Buccleuch Place \\ Edinburgh  EH8 9LW, Scotland\\ J.Hitzeman@ed.ac.uk}

\title{Algorithms for Analysing the Temporal Structure of 
Discourse\thanks{~We
would like to thank Alex Lascarides and Massimo Poesio for comments on an
earlier draft.}~\thanks{~This work was supported in 
part by the European Commission's programme on
{\it Linguistic Research and Engineering} through
project LRE-61-062, ``Towards a declarative theory of discourse.''}}

\twocolumn
\newcommand{\ignore}[1]{}

\newcommand{\brule}{\small \begin{verbatim}}
\newcommand{\erule}{\end{verbatim} \normalsize}

\newcommand{\dcu}{{\sc dcu}}
\newcommand{\dcus}{{\sc dcu}s}
\newcommand{\hpsg}{{\sc hpsg}}
\newcommand{\ale}{{\sc ale}}

\begin{document}
\maketitle

\begin{abstract}
We describe a method for analysing the temporal structure of
a discourse which takes into account the effects of tense, aspect,
temporal adverbials and rhetorical structure and which minimises
unnecessary ambiguity in the temporal structure.  It is part of a
discourse grammar
implemented in Carpenter's {\sc ale} formalism.
The method for building up the temporal structure of the discourse
combines constraints and preferences: we use constraints to reduce the
number of possible structures, exploiting the {\sc hpsg} type hierarchy and
unification for this purpose; and we apply preferences to choose
between the remaining options using a temporal centering mechanism.
We end by recommending that an underspecified representation of the structure
using these techniques be used to avoid generating the temporal/rhetorical
structure until higher-level information can be used to disambiguate.
\end{abstract}

%********************************************************
\section{Introduction}
In this paper we describe a method for analysing the temporal
structure of a discourse. This component was implemented as part of a
discourse grammar for English. The goals of the temporal component
were to yield a detailed representation of the temporal structure of
the discourse, taking into account the effect of tense, aspect and
temporal expressions while at the same time minimising unnecessary
ambiguity in the temporal structure.  The method combines a
constraint-based approach with an approach based on preferences: we
exploit the {\sc hpsg} type hierarchy and unification to arrive at a
temporal structure using constraints placed on that structure by
tense, aspect, rhetorical structure and temporal expressions, and we
use the {\it temporal centering} preferences described
by~\cite{Kameyama-et-al:ACL93,poesio:thesis} to rate the 
possibilities for temporal
structure and choose the best among them.

The starting point for this work was Scha and
Polanyi's discourse grammar (Scha \& Polanyi 1988;
Pr\"{u}st {\it et al} 1994).  For the
implementation we extended the {\sc hpsg}
grammar~\cite{pollard&sag:hpsg} which Gerald Penn and Bob Carpenter
first encoded in {\sc ale}~\cite{carpenter:ale-manual}.  This
paper will focus on our temporal processing algorithm,
and in particular on our analysis of narrative
progression, rhetorical structure, perfects and temporal expressions.

\section{Constraints on narrative continuations}

Probably the best known algorithm for tracking narrative progression
is that developed by Kamp~\shortcite{kamp:79}, 
Hinrichs~\shortcite{hinrichs:81}, and 
Partee~\shortcite{partee:84}, which formalises the
observation that an event will occur {\it just after} a preceding
event, while a state will {\it overlap} with a preceding event.  This
algorithm gives the correct results in examples such as the
following:
\vspace{.1in}
\begin{tabbing}
\NUM{jjk} John entered the room.  Mary stood up.\\
\NUM{jjk2} \parbox[t]{2.4in}{John entered the room.  
Mary was seated behind the desk.}
\end{tabbing}
\vspace{.1in}
In \sref{jjk} the event of Mary's standing is understood to occur just
after John enters the room, while in \sref{jjk2} the state in which
Mary is seated is understood to overlap with the event of John's
entering the room.

However, if there is a rhetorical
relationship between two eventualities such as causation, elaboration
or enablement, the temporal defaults can be overridden, as in the
following examples:
\vspace{.1in}
\begin{tabbing}
\NUM{mary} \=a. John fell. Mary pushed him.\\
\>b. \parbox[t]{2.22in}{Local builders constructed the Ford 
St. Bridge.  They used 3 tons of bricks.}
\end{tabbing}
\vspace{.1in}
In (\ref{mary}a) there is a causal relationship between Mary's pushing
John and his falling, and the second event is understood to precede
the first.  In (\ref{mary}b), the second sentence is an elaboration of
the first, and they therefore refer to aspects of the same event 
rather than to two sequential events.

It has been suggested that only world knowledge allows one to detect
that the default is being overridden here. For example, Lascarides \&
Asher (1991) suggest that general knowledge postulates (in the case of
(\ref{mary}a): that a pushing can cause a falling) can be invoked to
generate the backward movement reading.

The problem for practical systems is twofold:  we could assume that
in the case of narrative the Kamp/Hinrichs/Partee algorithm is the 
default, but
each time the default is applied we would need to check all our 
available
world knowledge to see whether there isn't a world knowledge
postulate which might be overriding this assumption. 
Clearly this would make the processing of text a very expensive
operation.

An alternative is to assume that the temporal ordering between events
in two consecutive sentences can be any of the four
possibilities ({\it just\_after}, {\it precede}, {\it same-event}
and {\it overlap}).  But then the resulting temporal structures will
be highly ambiguous even in small discourses. And sometimes this
ambiguity is unwarranted. Consider:
\vspace{.1in}
\begin{tabbing}
\NUM{marc} \parbox[t]{2.4in}{Mary stared at John.  He gave her back her slice
of pizza.}
\end{tabbing}
\vspace{.1in}
Here, it would appear, only one reading is possible, i.e.\ the one
where John gave Mary her slice of pizza {\it just\/} {\it after\/} 
she stared
or started to stare at him. 
It would be undesirable 
for the temporal processing mechanism to postulate an ambiguity
in this case.

Of course, sometimes it is possible to take advantage of certain cue
words which either indicate or constrain the rhetorical relation.  For
example, in \sref{4} the order of the events is understood to be the
reverse of that in \sref{jjk} due to the cue word {\it because\/}
which signals a causal relationship between the events:
\vspace{.1in}
\begin{tabbing}
\NUM{4} \parbox[t]{2.4in}{John entered the room because Mary stood up.}
\end{tabbing}
\vspace{.1in}
As Kehler~\shortcite{kehler:ACL94b} points out,
if forward movement of time is considered a default with
consecutive event sentences, then the use of ``because'' in \sref{4}
should cause a temporal clash---whereas it is perfectly
felicitous. 
Temporal expressions such as {\it at noon} and {\it the previous
Thursday} can have a similar effect: they too can override the default
temporal relations and place constraints on tense.  In \sref{cons1},
for example, the default interpretation would be that John's being in
Detroit overlaps with his being in Boston, but the phrase {\it the
previous Thursday} overrides this, giving the interpretation that
John's being in Detroit precedes his being in Boston:
\vspace{.1in}
\begin{tabbing}
\NUM{cons1} \parbox[t]{2.4in}{John was in Boston.  The previous 
Thursday he was in Detroit.}
\end{tabbing}
\vspace{.1in}
This suggests that 
the temporal information given by
tense acts as a weaker constraint on temporal structure than the 
information
given by temporal adverbials.  

The possibilities for rhetorical relations (e.g., whether something is
narration, or elaboration, or a causal relation) can be further 
constrained by
aspect. For example, a state can elaborate another state or an event:
\vspace{.1in}
\begin{tabbing}
\NUM{5} \=a. Mary was tired.  She was exhausted.\\
\>b. \parbox[t]{2.22in}{Mary built a dog house.  It was a labour of 
love.}
\end{tabbing}
But an event can only elaborate another event, as in \sref{alab}:

\begin{figure*}[htb]
\footnotesize
\begin{center}
\begin{tabular}{|l|c|p{9cm}|}
\multicolumn{3}{c}{Table~1.~Possible~relations~when~S$_{2}$~expresses
a~simple~past~event.}\\[2ex]
\hline
\multicolumn{1}{|c|}{{\bf S$_{1}$}} & {\bf Relation} & {\bf Example} 
\\ \hline\hline
 &  just-after S$_{1}$ & Mary pushed John. He fell. \\
past event  & precede S$_{1}$& John fell.  Mary pushed him. \\
 &  overlap S$_{1}$& {\sc no} \\
 &  same-event S$_{1}$& I assembled the desk myself. The drawers only took
me ten minutes. \\ \hline

 &  just-after S$_{1}$& Mary stared at John. He gave her back her slice 
of pizza. \\
past activity &  precede S$_{1}$& {\sc no} \\
 &   overlap S$_{1}$& {\sc no} \\
 &  same-event S$_{1}$& {\sc no} \\ \hline

 &  just-after S$_{1}$& {\sc no} \\
 &  just-after {\sc tf}$_{1}$& Sam arrived at eight.  He was tired.  
He rang
the bell.\\
 &  precede S$_{1}$& {\sc no} \\
past state  & precede {\sc tf}$_{1}$& ?John fell.  He was in pain.
Mary pushed him.\\
 &   overlap S$_{1}$& Mary was angry. She pushed John. \\
 &   overlap {\sc tf}$_{1}$& {\sc no}\\
 &  same-event S$_{1}$& {\sc no} \\ 
 &  same-event {\sc tf}$_{1}$& I assembled the desk myself.  It was 
beautiful.
The drawers
only took me ten minutes.\\ \hline

 &  just-after S$_{1}$& Sam had arrived at the house. He rang the 
bell. \\
 past perf event  & precede S$_{1}$& Sam arrived at the house. He had 
lost the key. He rang the bell. \\
 &  overlap S$_{1}$& {\sc no} \\
 &  same-event S$_{1}$& I had assembled the desk myself. The drawers 
only took
me ten minutes. \\ \hline

 &  just-after S$_{1}$& Mary had stared at John. He gave her back her 
slice of pizza. \\
past perf activity   & precede S$_{1}$& {\sc no} \\
 &  overlap S$_{1}$& {\sc no} \\
 &  same-event S$_{1}$&{\sc no} \\ \hline

 &  just-after S$_{1}$& {\sc no} \\
 &  just-after {\sc tf}$_{1}$& Martha discovered the broken lock.  
Someone had
been in the garage.  They rearranged the tools.\\
 &  precede S$_{1}$& {\sc no} \\
past perf state   & precede {\sc tf}$_{1}$& {\sc no}\\
 &  overlap S$_{1}$& Martha discovered the broken lock.
Someone had been in the garage. They rearranged the tools. \\
 &  overlap {\sc tf}$_{1}$& {\sc no}\\
 &  same-event S$_{1}$& {\sc no} \\ 
 &  same-event {\sc tf}$_{1}$& Mary built the desk herself.  She 
had been happy
taking it on.  The drawers only took her ten minutes.\\ \hline \hline 
\end{tabular}
\end{center} \normalsize
\end{figure*}
\begin{tabbing}
\NUM{alab} \=a. \parbox[t]{2.22in}{Mary built a dog house.  She used 
two tons of bricks.}\\[1ex]
\>b. \parbox[t]{2.22in}{Mary was tired/working hard.  ?She built a 
dog house.}
\end{tabbing}
\vspace{.1in}
For the eventive second sentence of (\ref{alab}b) to be an elaboration
of the first sentence, it must occur in a stative form---for example
as a progressive (i.e., {\it She was building a dog house}).

Because of considerations like these, our aim in the implementation
work was to treat tense, aspect, cue words and rhetorical relations as
mutually constraining, with more specific information such as explicit 
cue
words having higher
priority than less specific information such as tense.
The main advantage of
this approach is
that it 
reduces 
temporal structure ambiguity 
%\hspace{1in}
%\begin{picture}(188,442)(0,0)
%\end{picture}   
without having
to rely on detailed world knowledge postulates.

Table 1 lists the possible temporal relations between the
eventualities described by two consecutive sentences without temporal
expressions or cue words, where the first
sentence (S$_{1}$) may have any tense and aspect and the second
sentence (S$_{2}$) expresses a simple past event.  We constrain
S$_{2}$ in this way because of lack of space; additional constraints
are given in~\cite{hitzeman:discourse-d2}.
For example, if a simple past
eventive sentence follows a simple past eventive sentence the second
event can be 
understood to occur just after the first, to precede the
first or to refer to the same event as the first (an elaboration
relation), but the two events cannot overlap; these constraints are 
weaker,
however, than explicit clues such as cue words to rhetorical relations 
and
temporal expressions.
When S$_{1}$ expresses a state, it is possible for the temporal relation to
hold between the event described by S$_{2}$ and the event or activity 
most
closely preceding S$_{1}$, i.e., the temporal focus of S$_{1}$, here 
referred to as
{\sc tf}$_{1}$.\footnote{In this chart it appears
that whether the tense is simple past or past perfect makes no 
difference, and
that only aspect affects the possible temporal relations between 
S$_{1}$ and
S$_{2}$.  However, it is important not to ignore tense because other 
combinations of tense and aspect do show that tense
affects which relations are possible, e.g., a simple past stative 
S$_{2}$
cannot have a {\it precede\/} relation with any S$_{1}$, while a past 
perfect
stative S$_{2}$ can.}

However, we haven't solved the problem completely at this point:
although tense can provide 
a further constraint on the temporal structure of such discourses,
it can also add a further ambiguity.  Consider \sref{pp}:
\vspace{.06in}
\begin{tabbing}
\NUM{pp} Sam rang the bell.  He had lost the key.
\end{tabbing}
\vspace{.06in}
Clearly, the event described by the past perfect sentence must precede
the event described by the first, simple past sentence.
However, if a third sentence is added, an ambiguity results.
Consider the following possible continuations of \sref{pp}:
\vspace{.06in}
\begin{tabbing}
\NUM{pp2} \=a. ...Hannah opened the door.\\
\>b. \parbox[t]{2.15in}{...It fell through a hole in his pocket.}
\end{tabbing}
\vspace{.06in}
The temporal relation between these continuations and the portion of
earlier text they attach to is constrained along the lines sketched
before. The problem here is determining which thread in \sref{pp} they 
continue; (\ref{pp2}a) continues the thread in which Sam
rings the bell, but (\ref{pp2}b) continues the thread in which Sam
loses the key.

A further ambiguity is that when the third sentence is past perfect,
it may be a continuation of a preceding thread or the start of
a new thread itself. Consider:
\vspace{.06in}
\begin{tabbing}
\NUM{9} \=a. \parbox[t]{2.15in}{Sam rang the bell.  He had lost the key.  
It had fallen
through a hole in his pocket.}\\[1ex]
\>b. \parbox[t]{2.15in}{John got to work late. He had
left the house at 8. He had eaten a big breakfast.}
\end{tabbing}
\vspace{.06in}
In (a) the third sentence continues the thread about losing the key;
in (b) the third starts a new thread.\footnote{We will not discuss the
additional problem that if the final sentence in (\ref{9}b) is the end
of the text, the text is probably ill-formed.  This is because a
well-formed text should not leave threads ``dangling'' or unfinished.
This is probably also the reason for the awkwardness of the
well-known example {\it Max poured a cup of coffee. He had entered the
room.}}

For the problem with multi-sentence discourses, and the ``threads''
that sentences continue, we use an implementation of temporal
centering~\cite{Kameyama-et-al:ACL93,poesio:thesis}.  This is a 
technique similar to
the type of centering used for nominal
anaphora~\cite{sidner:83,GJW:83}.  Centering assumes that discourse
understanding requires some notion of ``aboutness.''  While nominal
centering assumes there is one object that the current discourse is
``about,'' temporal centering assumes that there is one thread that
the discourse is currently following, and that, in addition to tense
and aspect constraints, there is a preference for a new utterance to
continue a thread which has a parallel tense or 
which is semantically related to it and a preference
to continue the current thread rather than switching to another
thread.  Kameyama {\it et al.} (1993) confirmed these preferences when
testing their ideas on the Brown corpus.

\begin{figure*}[htb]
\epsfbox{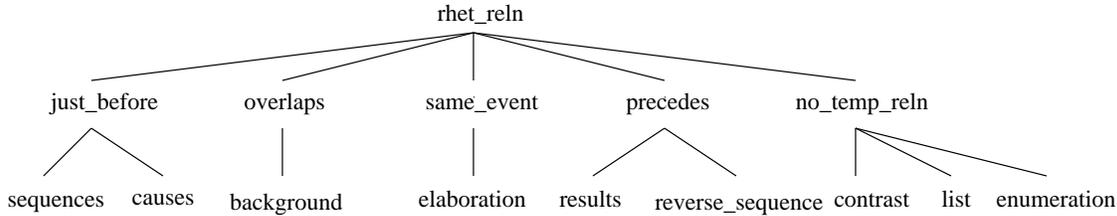}
\caption{The type hierarchy used for constraints}
\end{figure*}

As an example of how the temporal centering preference techniques can 
reduce
ambiguity, recall example \sref{pp} and the possible continuations shown in
\sref{pp2}.
The difficulty in these examples is determining whether the third
sentence continues the thread begun by the first or second sentence.
For example, in (\ref{pp2}a) 
the preference technique which
allows us to choose the first thread over the second is one which 
assigns a
higher rating to a thread whose tense is parallel to that of
the new
sentence; in this case both {\it Sam rang the bell} and {\it Hannah 
opened the
door} are in the simple past tense.
In example (\ref{pp2}b) the fact that the key is mentioned only in the
second sentence of \sref{pp} links (\ref{pp2}b) with the second thread.
To handle an example like \sref{third}, we employ a preference for 
relating a
sentence to a thread that has content words that are rated as 
semantically
``close'' to that of the sentence:
\begin{tabbing}
\NUM{third} \parbox[t]{2.35in}{Sam rang the bell.  He had lost the key.  
His
keyring broke.}
\end{tabbing}

We store semantic patterns between words as a
cheap and quick form of world knowledge;
these patterns are easier 
to provide than are the detailed
world knowledge postulates required in some other approaches, and result
in similar and sometimes more precise temporal structures with less
processing overhead.
Using the semantic patterns we know that {\it key\/} and {\it keyring\/} are
semantically close, and through \linebreak
that semantic link between the second and third
sentences we prefer to connect the third sentence to the thread begun
by the second.\footnote{Semantic closeness ratings won't help in examples
\sref{pp} -- \sref{pp2} because
there is as strong a relationship between {\it door\/} and {\it bell\/} as
there is between {\it door\/} and {\it key\/}.}  The approach to 
representing semantic relationships we
take is one used by Morris \& Hirst~\shortcite{morris(jane)-hirst91}
wherein the words in the lexicon are associated with each other
in a thesaurus-like fashion
and given a rating according to how semantically ``close'' they are.
We thus avoid relying on high-level inferences and very specific world
knowledge postulates, our goal being to determine the temporal
structure as much as possible prior to the application of higher-level
inferences.

\section{An {\sc hpsg} implementation of a discourse grammar}

Following Scha \& Polanyi~\shortcite{scha&polanyi:88} and Pr\"{u}st 
{\it et al} (1994), our model of discourse
consists of units called Discourse Constituent Units (\dcus) which are
related by various temporal and rhetorical relations.  A basic \dcu\ 
represents a
sentence (or clause), and complex \dcus\ are built up from basic and complex
\dcus.  \ignore{
In our \hpsg\ implementation, a \dcu\ is simply a sign with certain
information that is unnecessary for discourse processing removed.}

In our {\sc ale} implementation, a \dcu\ contains the following slots
for temporal information:

\begin{description}
\item[{\sc cue\_word:}] Cues to rhetorical structure, e.g., ``because.''
\item[{\sc v\_and\_np\_list:}]  Contains content words found in this 
{\sc dcu},
and is used to compare the content words  of  the  current  \dcu\  with 
those
in previous threads, in order to rate the semantic ``closeness'' of the 
\dcu\
to each thread.
\item[{\sc sem\_aspect:}] Contains the semantic aspect ({\em event, state,
activity}).  We have extended the Penn \& Carpenter implementation of the 
\hpsg\ grammar so that semantic aspect is calculated
compositionally (and stored here).
\item[{\sc rhet\_reln:}] The relation between this {\sc dcu} and a previous 
one. Lexical items and phrases such as cue words (stored in {\sc cue\_word})
affect the value of this slot.
\item[{\sc temp\_center:}]      Used for temporal centering; Keeps track of 
the thread currently being followed (since
there is a preference for continuing the current thread) and all the threads
that have been constructed so far in the discourse.
\begin{description}
  \item[{\sc fwd\_center:}]  Existing threads
  \item[{\sc bkwd\_center:}] The thread currently being followed
  \item[{\sc closed\_threads:}]~Threads no longer available for continuation
\end{description}
\item[{\sc temp\_expr\_relns:}]   Stores the semantic interpretation of 
temporal expressions associated with this
\dcu.
\item[{\sc temp\_relns:}]       Stores the temporal relations between the
eventualities in the discourse. 
\item[{\sc tempfoc:}]      The most recent event in the current thread 
which a
subsequent eventuality may elaborate upon ({\it same-event}),
{\it overlap}, come {\it just\_after} or
{\it precede}. 
\item[{\sc tenasp:}]     Keeps track of the tense and syntactic aspect 
of the \dcu\
(if the \dcu\ is simple).
\begin{description}
  \item[{\sc tense:}] past, pres, fut
  \item[{\sc aspect:}]  simple, perf, prog, perf\_prog
\end{description}
\end{description}

To allow the above-mentioned types of information to mutually constrain
each other, we employ a hierarchy of rhetorical and temporal relations
(illustrated in Figure 1), using the {\sc ale} system in such a way
that clues such as tense and cue words work together to reduce the
number of possible temporal structures.  This approach improves upon
earlier work on discourse structure such
as~\cite{lascarides&asher:ACL91} and~\cite{kehler:ACL94b} in reducing
the number of possible ambiguities; it is also more precise than
the Kamp/Hinrichs/Partee approach in that it takes into account ways in 
which the
apparent defaults can be overridden and differentiates between events and
activities, which behave differently in narrative progression.

Tense, aspect, rhetorical relations and temporal expressions affect the 
value of the {\sc
rhet\_reln} type that expresses 
the relationship between two \dcus:
cue words are lexically marked according to what rhetorical relation
they specify, and this relation is passed on to the \dcu.  
Explicit relation markers such as cue words and temporal relations must be
consistent and take priority over indicators such as tense and aspect.  For
example, sentence \sref{ruled} will be ruled out because the cue phrase 
{\it as
a result} conflicts with the temporal expression {\it ten minutes earlier}:
\vspace{.1in}
\begin{tabbing}
\NUM{ruled} \parbox[t]{2.35in}{\#Mary pushed John and as a result ten 
minutes earlier he fell.}
\end{tabbing}
\vspace{.1in}
On the other hand, if temporal expressions indicate an overlap relation 
and cue
words indicate a background relation as in \sref{cl}, these contributions 
are consistent and
the {\sc rhet\_reln} type will contain a {\it background\/} value (the more
specific value of the two):
\begin{tabbing}
\NUM{cl} \parbox[t]{2.3in}{Superman stopped the train just in time.  
Meanwhile, Jimmy Olsen was in trouble.}
\end{tabbing}

\section{The algorithm}

For reasons of space it is difficult to give examples of the
sign-based output of the grammar, or of the \ale\ rules, so we will
restrict ourselves here to a summary of the algorithm and to a very
limited rendition of the system output. The algorithm used for
calculating the temporal structure of a discourse can be summarised as
follows. It consists of two parts, the constraint-based portion and
the preference-based portion:

\begin{enumerate}
  \item The possible temporal/rhetorical relations are constrained.
\begin{enumerate}
  \item If there is a temporal expression, it determines the temporal
relationship of the new \dcu\ to the previous ones, and defaults are ignored.
  \item Lexical items such as cue words influence the value of the {\sc
rhet\_reln} type (See Figure 1).
  \item If steps (a) and (b) attempt to place conflicting values in the {\sc
rhet\_reln} slot, the parse will fail.
  \item If there is no temporal expression or cue phrase, tense and semantic 
aspect also influence the value of the {\sc
rhet\_reln} type (See Table 1), so that rhetorical relations, tense and 
aspect constrain each
other.  
\end{enumerate}
  \item If more than one possibility exists, semantic preferences are used to
choose between the possibilities.
\begin{enumerate}
  \item A ``semantic distance'' rating between the new \dcu\ and each previous
thread is determined.  (If there are no existing threads a new thread is
started.)
  \item Other preferences, such as a preference for relating the new 
\dcu\ to a
thread with parallel tense, are employed
(See~\cite{Kameyama-et-al:ACL93,poesio:thesis} for details), and the 
resulting ratings are
factored into the rating for each thread.
  \item If the thread currently being followed is among the highest rated
threads, this thread is continued.  (This corresponds to temporal centering's
preference to continue the current thread.)
  \item If not, the \dcu\ may continue any of the highest rated threads, and
each of these solutions is generated.
\end{enumerate}
\end{enumerate} %\dsp

Charts such as Table 1 provide the observations we use to fill in the 
value of
{\sc rhet\_reln}.  Those observations are summarised below. In
what follows, the event
variable associated with \dcu$_{i}$ is e$_{i}$ and the {\sc tempfoc} of 
e$_{1}$
is the most recent event/activity processed, possibly e$_{1}$ itself:

\begin{itemize} 
  \item e$_{2}$ can overlap with e$_{1}$ if 
\begin{itemize}
  \item \dcu$_{2}$ describes a state, or
  \item \dcu$_{1}$ describes a state and \dcu$_{2}$ describes an activity.
\end{itemize}
  \item e$_{2}$ can occur just-after the {\sc tempfoc} of e$_{1}$ if 
\begin{itemize}
  \item \dcu$_{2}$ describes a simple tense event, or
  \item \dcu$_{1}$ describes a complex tense clause and \dcu$_{2}$ 
describes a
complex tense event, or
  \item \dcu$_{1}$ describes an event and \dcu$_{2}$ describes an atelic 
or a
simple tense state, or
  \item \dcu$_{1}$ describes a state and \dcu$_{2}$ describes a simple tense 
activity.
\end{itemize}
  \item e$_{2}$ can precede e$_{1}$ if 
\begin{itemize}
  \item \dcu$_{2}$ describes an event, or
  \item \dcu$_{1}$ doesn't describe an activity and \dcu$_{2}$ describes 
a past perfect stative.
\end{itemize}
  \item e$_{2}$ can elaborate on e$_{1}$ if
\begin{itemize}
  \item \dcu$_{1}$ describes an event, or
  \item \dcu$_{1}$ describes an activity and \dcu$_{2}$ describes an 
atelic, or
  \item \dcu$_{1}$ and \dcu$_{2}$ describe states and either \dcu$_{2}$
describes a simple tense state or \dcu$_{1}$ describes a complex tense state.
\end{itemize}
\end{itemize}

Using this algorithm, we can precisely identify the rhetorical and temporal
relations when cue words to rhetorical structure are present, as in \sref{j1}:
\vspace{.1in}
\begin{tabbing} 
%\ssp
\NUM{j1} \=\parbox[t]{2.35in}{John fell (e$_{1}$) because Mary pushed him 
(e$_{2}$).}\\
\>{\sc temp\_relns:} e$_{2}$ precedes e$_{1}$
\end{tabbing} 
\vspace{.1in}
%\dsp
We can also narrow the possibilities when no cue word is present by using
constraints based on observations of tense and aspect interactions such as
those shown in Table 1.  For example, if \dcu$_{1}$ represents a simple 
past eventive
sentence and \dcu$_{2}$ a past perfect eventive sentence, then in spite 
of the lack of
rhetorical cues we know that e$_{2}$ precedes e$_{1}$, as in \sref{j2}:
\vspace{.1in}
\begin{tabbing} 
\NUM{j2} \=\parbox[t]{2.35in}{Sam rang the doorbell (e$_{1}$).  He had 
lost the key
(e$_{2}$).}\\
\>{\sc temp\_relns:} e$_{2}$ precedes e$_{1}$
\end{tabbing} 
\vspace{.1in}
%\dsp
Also, when several structures are possible we can narrow the possibilities by
using preferences, as in the examples below:
\vspace{.1in}
\begin{tabbing} %\ssp
\NUM{vvg} \=\parbox[t]{2.35in}{Sam arrived at the house at eight (e$_{1}$).  
He had lost the key (e$_{2}$).}\\
\>a. \=...He rang the bell (e$_{3}$).\\
\>\>{\sc temp\_relns:} \=e$_{2}$ precedes e$_{1}$, \\ 
\>\>\>e$_{3}$ just-after e$_{1}$\\
\>b. \parbox[t]{2.15in}{...It fell through a hole in his pocket (e$_{3'}$).}\\
\>\>{\sc temp\_relns:} \=e$_{2}$ precedes e$_{1}$, \\ 
\>\>\>e$_{3'}$ just-after e$_{2}$
\end{tabbing} 
\vspace{.1in}
%\dsp
If we allow any of the four possible temporal relations between events, both
continuations of sentence \sref{vvg} would have 17 readings (4 x 4 + 1 reading
in which the third sentence begins a new thread).
Using constraints, we reduce the number of readings to 4.  Using preferences,
we reduce that to 2 readings for each continuation.  The correct temporal
relations are shown in \sref{vvg}.\footnote{The other reading, in which 
the third
sentence is an elaboration of one of the preceding events, must not be ruled
out
because there are cases such as {\it Sam arrived at the house at eight.  He
rang the bell.  He let it ring for two minutes,} in which such 
elaboration is possible.}

\section{An underspecified representation}
By using constraints and preferences, we can considerably reduce the 
amount of
ambiguity in the temporal/rhetorical structure of a discourse.  However,
explicit cues to rhetorical and temporal relations are not always available,
and these cases result in more ambiguity than is desirable when processing
large discourses.

Consider, however, that instead of generating all the possible
temporal/rhetorical structures, we could use the information available to fill
in the most restrictive type possible in the type hierarchy of
temporal/rhetorical relations shown in Figure 1.  
We can then avoid generating the structures until higher-level information can
be applied to complete the disambiguation process.

\section{Conclusion}

We presented a brief description of an algorithm for determining the
temporal structure of discourse. The algorithm is part of an {\sc
hpsg}-style discourse grammar implemented in Carpenter's {\sc ale}
formalism. Its novel features are that it treats tense, aspect,
temporal adverbials and rhetorical relations as mutually constraining; it
postulates less ambiguity than current temporal structuring algorithms
do; and it uses semantic closeness and other preference techniques 
rather than full-fledged world
knowledge postulates to determine preferences over remaining
ambiguities.  We also recommended using an underspecified representation of
temporal/rhetorical structure to avoid generating all solutions until
higher-level knowledge can aid in reducing ambiguity.

\end{document}